\documentclass[10pt,twocolumn,letterpaper]{article}

\usepackage{cvpr}
\usepackage{times}
\usepackage{epsfig}
\usepackage{graphicx}
\usepackage{amsmath}
\usepackage{amssymb}
\usepackage[algo2e]{algorithm2e}
\usepackage{ulem}
\newcommand{\apbbox}[1]{AP$^\text{bb}_\text{#1}$}
\newcommand{\apmask}[1]{AP$^\text{mk}_\text{#1}$}
\usepackage{algorithm}
\usepackage{listings}
\usepackage{multirow}
\usepackage{makecell}
\usepackage{blindtext}
\usepackage[numbers]{natbib}
\usepackage[utf8]{inputenc} %
\usepackage{subfigure}
\usepackage{booktabs}
\usepackage{color, colortbl}
\definecolor{Gray}{gray}{0.9}
\definecolor{Goldenrod}{RGB}{245,245,220} %
\newcommand{\gain}[1]{\textcolor{Green}{(+{#1})}}
\newcommand{\loss}[1]{\textcolor{red}{(-{#1})}}
\usepackage[table,xcdraw,dvipsnames]{xcolor}

\newcommand{\tablestyle}[2]{\setlength{\tabcolsep}{#1}\renewcommand{\arraystretch}{#2}\centering\small}

\usepackage{amssymb}
\usepackage{pifont}
\usepackage[pagebackref=true,breaklinks=true,letterpaper=true,colorlinks,bookmarks=false]{hyperref}

\cvprfinalcopy 

\def\cvprPaperID{3562} 

\ifcvprfinal\fi

\begin{document}

\title{TransMix: Attend to Mix for Vision Transformers}

\author{
	Jie-Neng Chen$^{1}$\footnotemark[1]
	\;\; Shuyang Sun$^{2}$\footnotemark[1]
	\;\; Ju He$^1$
	\;\; Philip Torr$^2$
	\;\; Alan Yuille$^1$
	\;\; Song Bai$^3$ \\
	$^1$Johns Hopkins University \;\; $^2$University of Oxford \;\;  $^3$ByteDance Inc. 
}

\maketitle
\renewcommand{\thefootnote}{\fnsymbol{footnote}}
\footnotetext[1]{These authors contributed equally to this work. \\
Correspondence to Jie-Neng Chen \href{jienengchen01@gmail.com}{(jienengchen01@gmail.com)} and Shuyang Sun \href{kevinsun@robots.ox.ac.uk}{(kevinsun@robots.ox.ac.uk)}}


\begin{abstract}
Mixup-based augmentation has been found to be effective for generalizing models during training, especially for Vision Transformers (ViTs) since they can easily overfit. However, previous mixup-based methods have an underlying prior knowledge that the linearly interpolated ratio of targets should be kept the same as the ratio proposed in input interpolation. This may lead to a strange phenomenon that sometimes there is no valid object in the mixed image due to the random process in augmentation but there is still response in the label space. To bridge such gap between the input and label spaces, we propose TransMix, which mixes labels based on the attention maps of Vision Transformers. The confidence of the label will be larger if the corresponding input image is weighted higher by the attention map.
TransMix is embarrassingly simple and can be implemented in just a few lines of code without introducing any extra parameters and FLOPs to ViT-based models. Experimental results show that our method can consistently improve various ViT-based models at scales on ImageNet classification. After pre-trained with TransMix on ImageNet, the ViT-based models also demonstrate better transferability to semantic segmentation, object detection and instance segmentation. TransMix also exhibits to be more robust when evaluating on 4 different benchmarks. 
Code will be made publicly available at \href{https://github.com/Beckschen/TransMix}{https://github.com/Beckschen/TransMix}.

\end{abstract}

\section{Introduction}
Transformers~\cite{transformer} have been dominant in nearly all tasks in natural language processing. Recently, transformer-based architectures like Vision Transformer (ViT) \cite{dosovitskiy2020image} have been introduced into the field of computer vision and show great promise on tasks like image classification~\cite{dosovitskiy2020image, el2021xcit, liu2021swin, touvron2021going}, object detection~\cite{wang2021pyramid, liu2021swin, YOLOS} and image segmentation~\cite{wang2021pyramid, liu2021swin, strudel2021}. However, recent works have found that ViT-based networks are hard to optimize and can easily overfit if the training data is not sufficient. A quick solution to this problem is to apply data augmentation and regularization techniques during training. Among them, the mixup-based methods like Mixup \cite{zhang2017mixup} and CutMix \cite{yun2019cutmix} are proven to be particularly helpful for generalizing the ViT-based network \cite{touvron2021training}.

Mixup takes a pair of inputs $\mathbf{x}_A, \mathbf{x}_B$ and their corresponding labels $\mathbf{y}_A, \mathbf{y}_B$, then creates an artificial training example $\lambda \mathbf{x}_A + (1-\lambda) \mathbf{x}_B$ with $\lambda \mathbf{y}_A + (1-\lambda) \mathbf{y}_B$ as its ground truth. Here $\lambda \in [0, 1]$ is the random mixing proportion sampled from a Beta distribution. This pre-assumes that linear interpolations of feature vectors should lead to linear interpolations of the associated targets.

\begin{figure}[!tbp]
    \centering
    \includegraphics[scale=0.7]{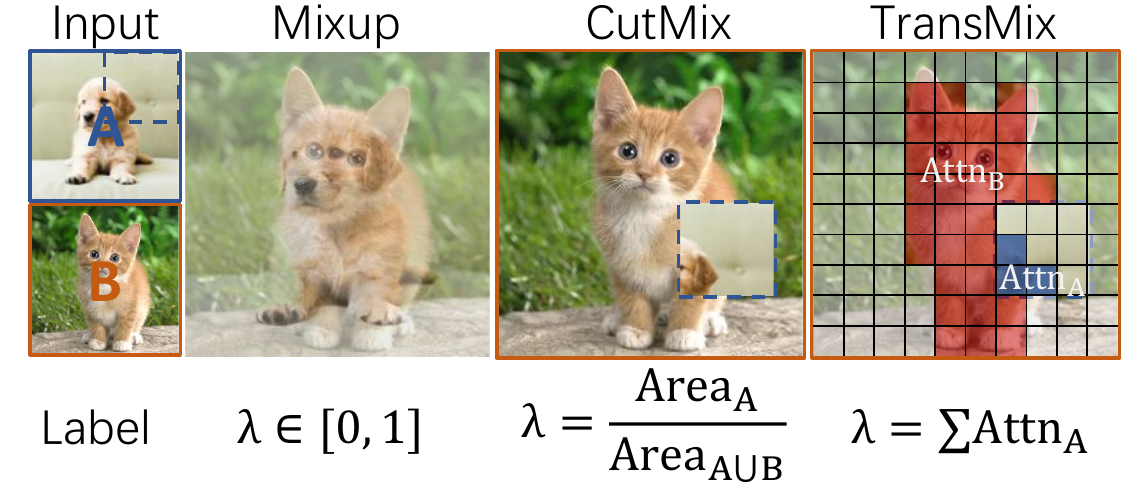}
    \caption{Mixup \cite{zhang2017mixup} and CutMix \cite{yun2019cutmix} samples $\lambda$ (proportion of label $\mathbf{y}_A$) randomly from a Beta distribution, while our TransMix calculates $\lambda$ with the sum of the values within the attention map that intersects with A (denoted as Attn$_\text{A}$, rendered in blue).}
    \label{fig:teaser}
\end{figure}

However, we argue that the above pre-assumption does not always stay true since \textbf{not all pixels are created equal}. As shown in Figure \ref{fig:teaser}, pixels in the background will not contribute to the label space as equally as those in the salient area. Some existing works \cite{walawalkar2020attentive, uddin2020saliencymix, kim2020puzzle} also find this problem and solve it by means of only mixing the most descriptive parts on the input level. Nevertheless, manipulating on inputs with the above methods may narrow the space of augmentation since they tend to less consider to put the background image into the mixture. Meanwhile, the above methods cost more number of parameters and/or training throughput to extract the salient region of input. For example, Puzzle-Mix~\cite{kim2020puzzle} requires model to forward and backward twice in an iteration and Attentive-Cutmix~\cite{walawalkar2020attentive} introduce a 24M external CNN to extract salient features.

Instead of investigating how to better mix images on the input level, in this paper, we focus more on how to mild the gap between the input and the label space through the learning of label assignment.
We find that the attention maps that are naturally generated in Vision Transformers can be well suited for this job. As shown in Figure \ref{fig:teaser}, we simply set $\lambda$ (weight of $\mathbf{y}_A$) as the sum of weights of attention map lying in $A$. 
In this way, the labels are re-weighted by the significance of each pixel instead of linearly interpolated with the same ratio as the mixed inputs.
Since the attention map is naturally generated in ViT-based models, our method can be merged into the their training pipeline with no extra parameters and minimal computation overhead.

\begin{figure}[!tbp]
\centering
\includegraphics[width=\linewidth]{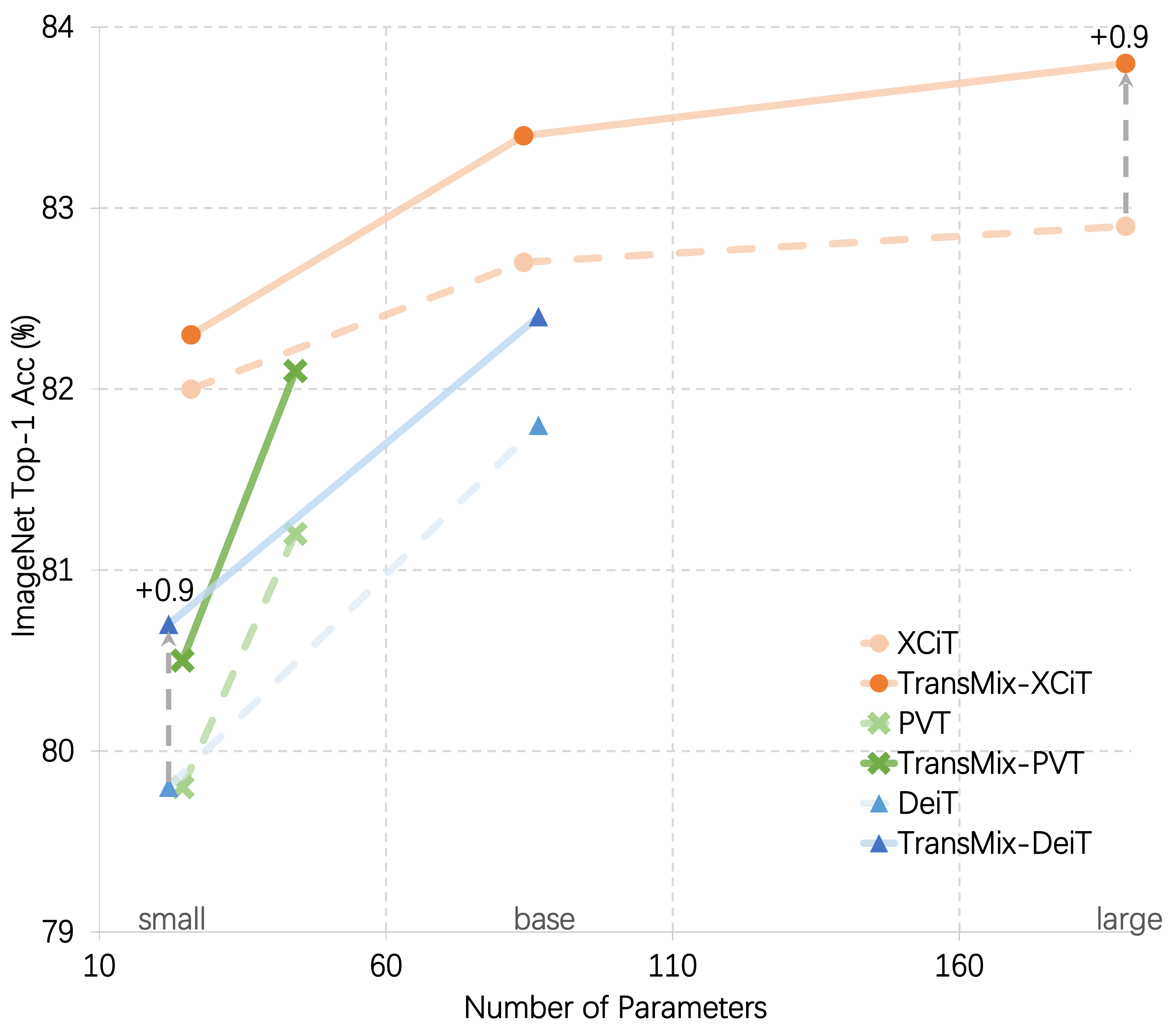}

\caption{TransMix can steadily improve a wide range of state-of-the-art ViT-based models on ImageNet with no parameter and minimal computation overhead. See results for more model variants in Table~\ref{tab:cls}.}
\label{fig:head}
\end{figure}

We show that such frustratingly simple idea can lead to consistent and remarkable improvement for a wide range of tasks and models. As exhibited in Fiugre \ref{fig:head}, TransMix can steadily boost all the listed ViT-based models. Notably, TransMix can further lift the top-1 accuracy on ImageNet by $0.9\%$ for both DeiT-S and a large variant XCiT-L. Interestingly, the largest model XCiT-L gains the most among all XCiT model scales.

Moreover, we demonstrate that if the model is first pre-trained with TransMix on ImageNet, the superiority can be further transferred onto downstream tasks including object detection, instance segmentation, semantic segmentation and weakly-supervised object segmentation/localization. We also observe that TransMix can help the model to be more rubust after evaluating it on 4 different benchmarks.

\section{Related Work}
\vspace{1ex}
\noindent\textbf{Vision Transformers (ViTs).}
Recently, Vision Transformer (ViT)~\cite{dosovitskiy2020image} was proposed to adapt the Transformer for image recognition by tokenizing and flattening images into a sequence of tokens. ViT is based on a sequence of Transformer blocks consisting of multi-head self attention layers and feed-forward networks. DeiT~\cite{touvron2021training} strengthens ViT by introducing a powerful training recipe and adopting knowledge distillation. Built upon the success of ViT, many efforts have been devoted to improving ViT and adapting it into various vision tasks including image classification~\cite{touvron2021training, touvron2021going, el2021xcit, yu2021glance, heo2021pit, liu2021swin, he2021transfg}, object localization/detection~\cite{Gao2021TSCAMTS, wang2021pyramid, liu2021swin, YOLOS} and image segmentation~\cite{wang2021pyramid, liu2021swin, strudel2021, chen2021transunet}.

\vspace{1ex}
\noindent\textbf{Mixup and its variants.}
Data augmentation has been widely studied to prevent DeepNets from over-fitting to the training data. To train and improve vision Transformer stably, Mixup and CutMix are two of the most helpful augmentation methods~\cite{touvron2021training}.  Mixup~\cite{zhang2017mixup} is a successful image mixture technique that obtains an augmented image by pixel-wisely weighted combination of two global images. The following Mixup variants~\cite{verma2019manifold, shen2020mix, gong2021keepaugment, hendrycks2019augmix, yun2019cutmix, walawalkar2020attentive, uddin2020saliencymix, kim2020puzzle} can be categorized into global image mixture (e.g. Manifold-Mixup~\cite{verma2019manifold}, Un-Mix~\cite{shen2020mix}) and regional image mixture (e.g. CutMix~\cite{yun2019cutmix}, Puzzle-Mix~\cite{kim2020puzzle}, Attentive-CutMix~\cite{walawalkar2020attentive} and SaliencyMix~\cite{uddin2020saliencymix}). Among all Mixup variants, the saliency-based methods including the attentive-CutMix, puzzle-Mix and saliency-CutMix are the most similar ones to our approach. However, TransMix has two fundamental differences with them: (1) Previous saliency-based methods \eg \cite{kim2020puzzle, walawalkar2020attentive, uddin2020saliencymix} enforce the image patch cropped in a salient region of the input image. Instead of manipulating in the input space, our TransMix focuses on how to more accurately assigning labels in the label space. (2) Previous saliency-based methods like \cite{walawalkar2020attentive} may use extra parameters to extract the saliency region. TransMix naturally exploits the Transformer's attention mechanism without any extra parameters. Experimental results also show that TransMix can lead to better results on ImageNet compared with these methods.

\vspace{1ex}
\noindent\textbf{Data-adaptive loss weight assignment.} TransMix re-assigns the ground truth labels with attentional guidance, which is related to data-adaptive loss weight assignment. Some existing works have found that the attention-like information can help to alleviate the long-tail problems for tasks like instance segmentation \cite{wang2021seesaw}, image demosaicing \cite{patchnet} \etc.

\section{TransMix}

\subsection{Setup and Background}
\label{sec:setup}

\vspace{1ex}
\noindent\textbf{CutMix data augmentation}
CutMix is a simple data augmentation technique combining two input-label pairs $(\mathbf{x}_{A}, \mathbf{y}_{A})$ and $(\mathbf{x}_{B}, \mathbf{y}_{B})$ to augment a new training sample $(\Tilde{\mathbf{x}}, \Tilde{\mathbf{y}})$. Formulaically, 
\begin{align}
    \mathbf{\Tilde{x}} & = \mathbf{M} \odot \mathbf{x}_{A} + (\mathbf{1}- \mathbf{M}) \odot \mathbf{x}_{B},
    \label{eq:cutmix_x}
    \\
    \mathbf{\Tilde{y}} & = \lambda \mathbf{y}_A + (1-\lambda) \mathbf{y}_B,
    \label{eq:cutmix_y}
\end{align}
where $\mathbf{M} \in \{0,1\}^{HW}$ denotes a binary mask indicating where to drop out and fill in from two images, $\mathbf{1}$ is a binary mask filled with ones, and $\odot$ is element-wise multiplication.  $\lambda$ is the proportion of $\mathbf{y}_A$ in the mixed label.

During augmentation, a randomly sampled region in $\mathbf{x}_B$ is removed and filled in with the patch cropped from $A$ of $\mathbf{x}_A$, where the patch's bounding box coordinates are uniformly sampled as $(r_x,r_y,r_w,r_h)$. The mixed-target assignment factor $\lambda$ is equal to the cropped area ratio $\frac{r_w r_h}{WH}$.

\vspace{1ex}
\noindent\textbf{Self-attention}
Self-attention, as introduced by~\cite{vaswani2017attention}, operates on an input matrix $\textbf{x} \in \mathbb{R}^{N \times d}$, where $N$ is the number of tokens, each of dimensionality $d$. 
The input  $\mathbf{x}$ is linearly projected to queries, keys and values, using the weight matrices $\mathbf{w}_{q} \in \mathbb{R}^{d \times d_q}$, $\mathbf{w}_{k} \in \mathbb{R}^{d \times d_k}$ and $\mathbf{w}_{v} \in \mathbb{R}^{d \times d_v}$, such that $\mathbf{q}{=}\mathbf{x} \mathbf{w}_q $, $\mathbf{k}{=}\mathbf{x} \mathbf{w}_k $ and $\mathbf{v}{=}\mathbf{x} \mathbf{w}_v $, where $d_q\!=\!d_k$.
Queries and keys are used to compute an attention map 
$\mathcal{A}(\mathbf{q},\mathbf{k}) = \textit{Softmax}(\mathbf{q} \mathbf{k}^\top/\sqrt{d_k}) \in \mathbb{R}^{N \times N}$,  
and the output of the self-attention operation is defined as the weighted sum of $N$ token features in $\mathbf{v}$ with the weights corresponding to the attention map: 
$\text{Attention}(\mathbf{q}, \mathbf{k}, \mathbf{v}) = \mathcal{A}(\mathbf{q}, \mathbf{k}) \textbf{v}$. 
 Single-head self-attention can be extended to multi-head self-attention by linearly projecting the queries, keys and values $g$ times with different, learned linear projections to $d_k$, $d_k$ and $d_v$ dimensions, respectively.

\subsection{TransMix}
\label{TransMix}
We propose TransMix to assign mixup labels with the guidance of attention map, where the attention map is defined specifically as the \textbf{multi-head class attention} $\mathbf{A}$, which is calculated as a part of self-attention. In the classification task, a class token is a query $\textbf{q}$ whose corresponding keys $\textbf{k}$ are the all input tokens, and class attention $\mathbf{A}$ is the attention map from the class token to the input tokens, summarizing which input tokens are the most useful to the final classifier. We then propose to use the class attention $\mathbf{A}$ to mix labels. 

\vspace{1ex}
\noindent\textbf{Multi-head Class Attention}
Vision Transformers (ViTs)\cite{dosovitskiy2020image} divide and embed an image $\mathbf{x} \in \mathbb{R}^{3\times H \times W}$ to $p$ patch tokens $\mathbf{x}_{patches} \in \mathbb{R}^{p \times d} $, and aggregate the global information by a class token $\mathbf{x}_{cls}  \in \mathbb{R}^{1 \times d}$, where $d$ is the dimension of embedding. ViTs operate on the patch embedding $\mathbf{z} = [\mathbf{x}_{cls}, \mathbf{x}_{patches}] \in \mathbb{R}^{(1+p) \times d}$.

Given a Transformer with $g$ attention heads and input patch embedding $\mathbf{z}$, we parametrize the multi-head class-attention with 
projection matrices $\mathbf{w}_{q}$, $\mathbf{w}_{k} \in \mathbb{R}^{d \times d}$. The class attention for each head can be formulated as:
\begin{align}
\mathbf{q} &= \mathbf{x}_{cls} \cdot \mathbf{w}_{q},\\
\mathbf{k} &= \mathbf{z} \cdot \mathbf{w}_{k}, \\
\mathbf{A}' &=\textit{Softmax}(\mathbf{q} \cdot \mathbf{k}^\top/\sqrt{d/g}),\\
\mathbf{A} &= \{\mathbf{A}'_{0, i}, \mid i \in [1, p]\},
\label{equ:A}
\end{align}

where $\textbf{q} \cdot \textbf{k}^\top \in \mathbb{R}^{1 \times (1+p)}$ indicates the class token is a query whose corresponding keys are the all input tokens, and $\mathbf{A} \in [0, 1]^{p}$ is the attention map from the class token to the image patch tokens, summarizing which patches are the most useful to the final classifier. When there are multiple heads in the attention, we simply average across all attention heads to obtain $ \mathbf{A} \in [0, 1]^{p}$. In implementation, $\mathbf{A}$ in Eqn.~\eqref{equ:A} is available as an intermediate output from the last Transformer block without architecture modification.

\vspace{1ex}
\noindent\textbf{Mixing labels with the attention map $\mathbf{A}$}
We follow the process of input mixture proposed in CutMix, which is defined in Eqn.~\eqref{eq:cutmix_x}, then we re-calculate $\lambda$ (the proportion of $\mathbf{y}_A$ in Eqn.~\eqref{eq:cutmix_y}) with the guidance of the attention map $\mathbf{A}$:

\begin{align} 
\lambda &= \mathbf{A} \cdot \downarrow(\mathbf{M}).
\label{equ:matmul}
\end{align}
Here $\downarrow(\cdot)$ denotes the nearest-neighbor interpolation down-sampling that can transform the original $\mathbf{M}$ from $HW$ into $p$ pixels . Note that we omit the dimension unsqueezing in Eqn.~\eqref{equ:matmul} for simplicity.
In this way, the network can learn to re-assign the weight of labels for each data point dynamically based on their responses in the attention map. The input that is better focused by the attention map will be assigned with a higher value in the mixed label.

\subsection{Pseudo-code}
Algorithm \ref{alg:code} provides the pseudo-code of TransMix in a pytorch-like style. The clean pseudo-code shows that simply few lines of code can boost the performance in the plug-and-play manner.

\begin{algorithm}[t]
\caption{\small{Pseudocode of TransMix in a PyTorch-like style.}}
\label{alg:code}

\definecolor{codeblue}{rgb}{0.25,0.5,0.5}
\lstset{
  backgroundcolor=\color{white},
  basicstyle=\fontsize{7.2pt}{7.2pt}\ttfamily\selectfont,
  columns=fullflexible,
  breaklines=true,
  captionpos=b,
  commentstyle=\fontsize{7.2pt}{7.2pt}\color{codeblue},
  keywordstyle=\fontsize{7.2pt}{7.2pt},
}
\begin{lstlisting}[language=python]
# H, W: the height and width of the input image
# p: number of patches
# M: 0-initialized mask with shape (H,W) 
# downsample: downsample from length (H*W) to (p)
# (bx1, bx2, by1, by2): bounding box coordinate

for (x, y) in loader:  # load a minibatch with N pairs
    # CutMix image in a minibatch
    M[bx1:bx2, by1:by2] = 1
    x[:,:,M==1] = x.flip(0)[:,:,M==1]
    M = downsample(M.view(-1))
    
    # attention matrix A: (N, p)
    logits, A = model(x)
    
    # Mix labels with the attention map
    lam = matmul(A, M)
    y = (1-lam) * y + lam * y.flip(0)

    CrossEntropyLoss(logits, y).backward()

\end{lstlisting}
\end{algorithm}

\section{Experiments}
In this section, we mainly demonstrate the effectiveness, transferability, robustness, and generalizability of TransMix. We verify the effectiveness of TransMix on ImageNet-1k classification in Section~\ref{sec:cls} and the transferability onto downstream tasks including semantic segmentation, object detection, and instance segmentation in Section~\ref{sec:tra}. The robustness of TransMix is examined on 4 benchmarks in Section~\ref{sec:rob}. Interestingly, we discover the mutual effects of TransMix and attention in Section~\ref{sec:mutual}. We validate the generalizability to Swin Transformer which is lacking class-token in Section~\ref{sec:swin}. Lastly, TransMix is compared with the state-of-the-art Mixup augmentation variants in Section~\ref{sec:sota}.

\subsection{ImageNet Classification}
\label{sec:cls}
\vspace{1ex}
\noindent\textbf{Implementation Details}
We use ImageNet-1k~\cite{deng2009imagenet} to train and evaluate our methods for image classification. ImageNet-1k consists of 1.28M training images and 50k validation images, labeled across 1000 semantic categories. The implementation is based on the Timm~\cite{rw2019timm} library. Unless specified otherwise, we make minimal changes to hyperparameters compared to the DeiT~\cite{touvron2021training} training recipe. We examined various baseline vision Transformer models including DeiT~\cite{touvron2021training}, PVT~\cite{wang2021pyramid}, CaiT~\cite{touvron2021going}, and XCiT~\cite{el2021xcit}, and the training schemes will be slightly adjusted to the official papers' implementations.  

All Transformers are trained for 300 epochs expect that \citet{el2021xcit} and \citet{touvron2021going} report 400 epochs for XCiT and CaiT respectively. As deploying DeiT~\cite{touvron2021training} training scheme, all baselines have already contained the carefully tuned regularization methods including RandAug~\cite{cubuk2020randaugment}, Stochastic Depth~\cite{huang2016deep}, Mixup~\cite{zhang2017mixup} and CutMix~\cite{yun2019cutmix}. To ease implementation, TransMix shares the same cropped region with CutMix for the input, whereas the label assignment is the mean of both methods. We throw away repeated augment~\cite{hoffer2020augment} due to its negative effects examined in ~\cite{heo2021pit}. We set warmup epoch to 20 expect DeiT-B keeping 5. The accuracy of our baseline implementation fluctuates only by $\pm 0.1\%$ compared with results reported in DeiT~\cite{touvron2021training}. The attention map $\mathbf{A}$ in Eqn.~\ref{equ:A} can be obtained as an intermediate output from the multi-head self-attention layer of the last Transformer block.

\vspace{1ex}
\noindent\textbf{Results}
As shown in Table~\ref{tab:cls}, TransMix can steadily boost the top-1 accuracy on ImageNet for all the listed models. No matter how complex the model is, TransMix can always help to boost the baseline performance.  Note that these models are with a wide range of model complexities, and the baselines are all carefully tuned with various data augmentation techniques \eg RandAug \cite{cubuk2020randaugment}, Mixup \cite{zhang2017mixup} and CutMix \cite{yun2019cutmix}. To be specific, TransMix can promote the top-1 accuracy of the small variant DeiT-S by $0.9\%$. Benefit from the higher attention quality, TransMix can also lift the top-1 accuracy of the large model XCiT-L by a remarkable $0.9\%$. We emphasize that these systematic improvement with just a tiny tweak on data augmentation is significant when compared with the structural modification on models. For example, CrossViT-B~\cite{chen2021crossvit} only lifts the DeiT-B baseline result by 0.4\% with 20.9\% parameters overhead while TransMix leads to more improvement in a parameter-free style.
Particularly, TransMix consistently boosts the base/large variants in the range of 0.6\% to 0.9\%, which is more striking than engineering new architectures such as PiT-B~\cite{heo2021pit}, T2T-24~\cite{Yuan_2021_ICCV}, CrossViT-B~\cite{chen2021crossvit} with the gains of 0.2\%, 0.5\%, 0.4\% respectively.


\begin{table}[t]
\tablestyle{3.8pt}{1.3}
\centering
\begin{tabular}{ccccc}
\toprule[1.5pt]
Models            & Params & FLOPs & \begin{tabular}[c]{@{}c@{}}Top-1 Acc \\ (\%) \end{tabular}&  \cellcolor[HTML]{F5F5DC} \begin{tabular}[c]{@{}c@{}} +TransMix \\ Top-1 Acc (\%) \end{tabular}      \\ \hline
DeiT-T~\cite{touvron2021training}            & 5.7M     & 1.6G  & 72.2   & \cellcolor[HTML]{F5F5DC} \textbf{72.6}  \\
PVT-T~\cite{wang2021pyramid}            & 13.2M     & 1.9G  & 75.1 & \cellcolor[HTML]{F5F5DC} \textbf{75.5} \\
XCiT-T~\cite{el2021xcit}            & 12M    & 2.3G  & 79.4   & \cellcolor[HTML]{F5F5DC} \textbf{80.1}  \\
\hline
CaiT-XXS\cite{touvron2021going}          & 17.3M  & 3.8G  & 79.1   & \cellcolor[HTML]{F5F5DC} \textbf{79.8}   \\
DeiT-S~\cite{touvron2021training}            & 22.1M    & 4.7G  & 79.8   & \cellcolor[HTML]{F5F5DC} \textbf{80.7}  \\
PVT-S~\cite{wang2021pyramid}             & 24.5M    & 3.8G  & 79.8 & \cellcolor[HTML]{F5F5DC} \textbf{80.5}   \\
XCiT-S~\cite{el2021xcit}            & 26M    & 4.8G  & 82.0   & \cellcolor[HTML]{F5F5DC} \textbf{82.3} \\
\hline
PVT-M~\cite{wang2021pyramid}            & 44.2M    & 6.7G & 81.2   & \cellcolor[HTML]{F5F5DC} \textbf{82.1}    \\
PVT-L~\cite{wang2021pyramid}             & 61.4M    & 9.8G  & 81.7 & \cellcolor[HTML]{F5F5DC} \textbf{82.4}  \\
XCiT-M~\cite{el2021xcit}            & 84M    & 16.2G & 82.7   & \cellcolor[HTML]{F5F5DC} \textbf{83.4}\\
DeiT-B~\cite{touvron2021training}            & 86.6M  & 17.6G & 81.8  & \cellcolor[HTML]{F5F5DC} \textbf{82.4}   \\
\hline
XCiT-L            & 189M   & 36.1G & 82.9   & \cellcolor[HTML]{F5F5DC} \textbf{83.8}  \\ 
\bottomrule[1.5pt]
\end{tabular}
\vspace{3pt}
\caption{TransMix can steadily boost the a wide range of model variants \eg DeiT, PVT, CaiT, and XCiT on ImageNet-1k classification. Note that all the baselines have been already carefully tuned with extensive augmentation and regularization techniques \eg Mixup \cite{zhang2017mixup}, CutMix\cite{yun2019cutmix}, RandAug\cite{cubuk2020randaugment}, DropPath\cite{droppath} \etc.}
\label{tab:cls}
\end{table}

\subsection{Transfer to Downstream Tasks}
\label{sec:tra}
ImageNet pre-training is the de-facto standard practice for many visual recognition tasks~\cite{he2019rethinking}.  Before training for downstream tasks, the weights pre-trained on ImageNet is used to initialize the Transformer backbone. We demonstrate the transferability of our TransMix-based pre-trained models on the downstream task including semantic segmentation, object detection and instance segmentation, on which we observe the improvements over the vanilla pre-trained baselines.


\vspace{1ex}
\noindent\textbf{Semantic Segmentation}
In our experiments, the sequence of patch encoding $z_{patches} \in \mathbb{R}^{p \times d}$ is decoded to a segmentation map $s\in \mathbb{R}^{H \times W \times K}$ where K is the number of semantic classes. We adopt two convolution-free decoders: (1) Linear decoder (2) Segmenter decoder. The reason for adopting the Linear decoder is to preserve the pre-trained information to the greatest extent. For linear decoder, a point-wise linear layer on DeiT patch encoding $z_{patches} \in \mathbb{R}^{p \times d}$ is used to produce patch-level logits $z_{lin} \in \mathbb{R}^{p \times K}$, which are  reshaped and bilinearly upsampled to segmentation map $s$. The Segmenter~\cite{strudel2021} decoder is a Transformer-based decoder namely Mask Transformer introduced in ~\cite{strudel2021, wang2021max}. 

We train and evaluate the models on the Pascal Context ~\cite{mottaghi_cvpr14} dataset and report Intersection over Union (mIoU) averaged over all classes as the main metric. The training set contains 4998 images with 59 semantic classes plus a background class. The validation set contains 5105 images. The training scheme follows ~\cite{mottaghi_cvpr14} which is built on MMSegmentation~\cite{mmseg2020}. As a reference, the result of ResNet101-Deeplabv3+~\cite{chen2017deeplab, chen2018encoder} is reported in MMSegmentation~\cite{mmseg2020}. 

According to Table~\ref{tab:seg}, TransMix pre-trained DeiT-S-Linear and DeiT-S-Segmenter improve over the vanilla pre-trained baselines by 0.6\% and 0.9\% mIoU respectively. There are consistent improvements on multi-scale testing.

\begin{table}[t]
\tablestyle{1.8pt}{1.3}
\begin{tabular}{cccccc}
\toprule[1.5pt]
Backbone                & Decoder                    & \begin{tabular}[c]{@{}c@{}}TransMix- \\ pretrained\end{tabular} & mAcc & mIoU  & \multicolumn{1}{c}{\makecell[c]{mIoU \\(MS)}} \\ \hline
ResNet101~\cite{he2016deep}                 & Deeplabv3+~\cite{chen2018encoder}                 &                          & 57.4      & 47.3 & 48.5                          \\ \hline
\multirow{4}{*}{DeiT-S~\cite{touvron2021training}} & \multirow{2}{*}{Linear}    &                          & 59.4     & 49.1 & 49.6                          \\
                        &                            & \cellcolor[HTML]{F5F5DC}\checkmark              & \cellcolor[HTML]{F5F5DC}\textbf{60.2}      & \cellcolor[HTML]{F5F5DC}\textbf{49.7} & \cellcolor[HTML]{F5F5DC} \textbf{50.3}                         \\ \cline{2-6} 
                        
                        &   \multirow{2}{*}{Segmenter~\cite{strudel2021}} &                          &  60.4     &  49.7 & 50.5                          \\
                       
                        &                            & \cellcolor[HTML]{F5F5DC} \checkmark               & \cellcolor[HTML]{F5F5DC} \textbf{61.4}      & \cellcolor[HTML]{F5F5DC} \textbf{50.6} & \cellcolor[HTML]{F5F5DC} \textbf{51.2}                          \\ 
\bottomrule[1.5pt]
\end{tabular}
\vspace{2pt}
\caption{Overhead-free impact of TransMix on transferring to downstream \textbf{semantic segmentation} task on the Pascal Context~\cite{mottaghi_cvpr14} dataset. (MS) denotes multi-scale testing.}
\label{tab:seg}
\end{table}

\vspace{1ex}
\noindent\textbf{Object Detection and Instance Segmentation}
Object detection and instance segmentation experiments are conducted on COCO 2017. All models are trained on 118K images and evaluated 5K validation images. We study on PVT~\cite{wang2021pyramid} as the detection backbone since its pyramid features make it favorable to object detection. The weights pre-trained on ImageNet is used to initialize the PVT backbone. We train and evaluate Mask R-CNN detector with the PVT backbone initialized with either vanilla (CutMix) or TransMix pre-trained weights for both object detection and instance segmentation. Following PVT~\cite{wang2021pyramid}, we adopt 1× training schedule (\ie, 12 epochs) to train the detector on mmDetection~\cite{chen2019mmdetection} framework. Results for Mask R-CNN with ResNet backbone are reported in mmDetection~\cite{chen2019mmdetection} as references. As showed in Table~\ref{tab:det}, we find that without introducing extra parameter, the detector initialized with TransMix-pretrained backbone improves over CutMix-pretrained backbone by 0.5\% box AP and 0.6\% mask AP. Note that regularization-based pre-training for backbone has limited capability on improving downstream object detection. For instance, the recent Mixup variant SaliencyMix~\cite{uddin2020saliencymix} only improved 0.16\% box AP over CutMix-pretrained model on a smaller detection dataset. 

\begin{table}[t]
\tablestyle{1.2pt}{1.3}
\begin{tabular}{ccccc|ccc}
\toprule[1.5pt]
\multirow{2}{*}{Backbone} & \multirow{2}{*}{Params} & \multicolumn{3}{c|}{Object detection} & \multicolumn{3}{l}{Instance segmentation} \\ 
                          &                         & \apbbox{~} &  \apbbox{50} & \apbbox{75}  & \apmask{~} & \apmask{50} & \apmask{75} \\ \hline
ResNet50~\cite{he2017mask}                 & 44.2M                    & 38.0      & 58.6        & 41.4        & 34.4         & 57.1         & 36.7         \\
ResNet101~\cite{he2017mask}                 & 63.2M                    & 40.4      & 61.1        & 44.2        & 36.4         & 57.7         & 38.8         \\
                          \hline
PVT-S~\cite{wang2021pyramid}   & 44.1M                    & 40.4      & 62.9        & 43.8        & 37.8         & 60.1         & 40.3         \\
\rowcolor[HTML]{F5F5DC}
TransMix-PVT-S              & 44.1M                    & \textbf{40.9}      & \textbf{63.8}        & \textbf{44.0}          & \textbf{38.4}         & \textbf{60.7}         & \textbf{41.3}         \\
\bottomrule[1.5pt]
\end{tabular}
\vspace{2pt}
\caption{Overhead-free impact of TransMix on transferring to downstream \textbf{object detection and instance segmentation} using Mask R-CNN~\cite{he2017mask} with PVT~\cite{wang2021pyramid} backbone on COCO val2017. \apbbox{~} denotes bounding box AP for object detection and \apmask{~} denotes mask AP for instance segmentation.}
\label{tab:det}
\end{table}

\subsection{Robustness Analysis}
\label{sec:rob}

Recently the discussions regarding the robustness of vision Transformer are emerging~\cite{naseer2021intriguing, mao2021towards, bai2020vitsVScnns}. To verify if TranMix can improve ViT-based models' robustness and out-of-distribution performance, we evaluated our TransMix pre-trained models on four robustness scenarios including occlusion, spatial structure shuffling, natural adversarial example, and out-of-distribution detection.

\vspace{1ex}
\noindent\textbf{Robustness to Occlusion}
\citet{naseer2021intriguing} studies whether ViTs perform robustly in occluded scenarios, where some or most of the image content is missing. To be specific, vision Transformers divide an image into M=196 patches belonging to a 14x14 spatial grid; \ie an image of size 224×224×3 is split into 196 patches, each of size 16×16×3. Patch Dropping means replacing original image patches with blank 0-value patches. As an example, dropping 100 such patches from the input is equivalent to losing 51\% of the image content. Following \cite{naseer2021intriguing}, we showcase the classification accuracy on ImageNet-1k validation set with three dropping settings. (1) \emph{Random Patch Dropping}: A subset of M patches is randomly selected and dropped. (2) \emph{Salient (foreground) Patch Dropping}: This studies the robustness of ViTs against occlusions of highly salient regions. \citet{naseer2021intriguing} thresholds DINO's attention map to obtain salient patches, which are dropped by ratios. (3) \emph{Non-salient (background) Patch Dropping}: The least salient regions of an image are selected and dropped following the same approach as above.

\begin{figure*}[!tbp]
\centering
\subfigure{\includegraphics[width=0.32\linewidth,height=0.26\linewidth]{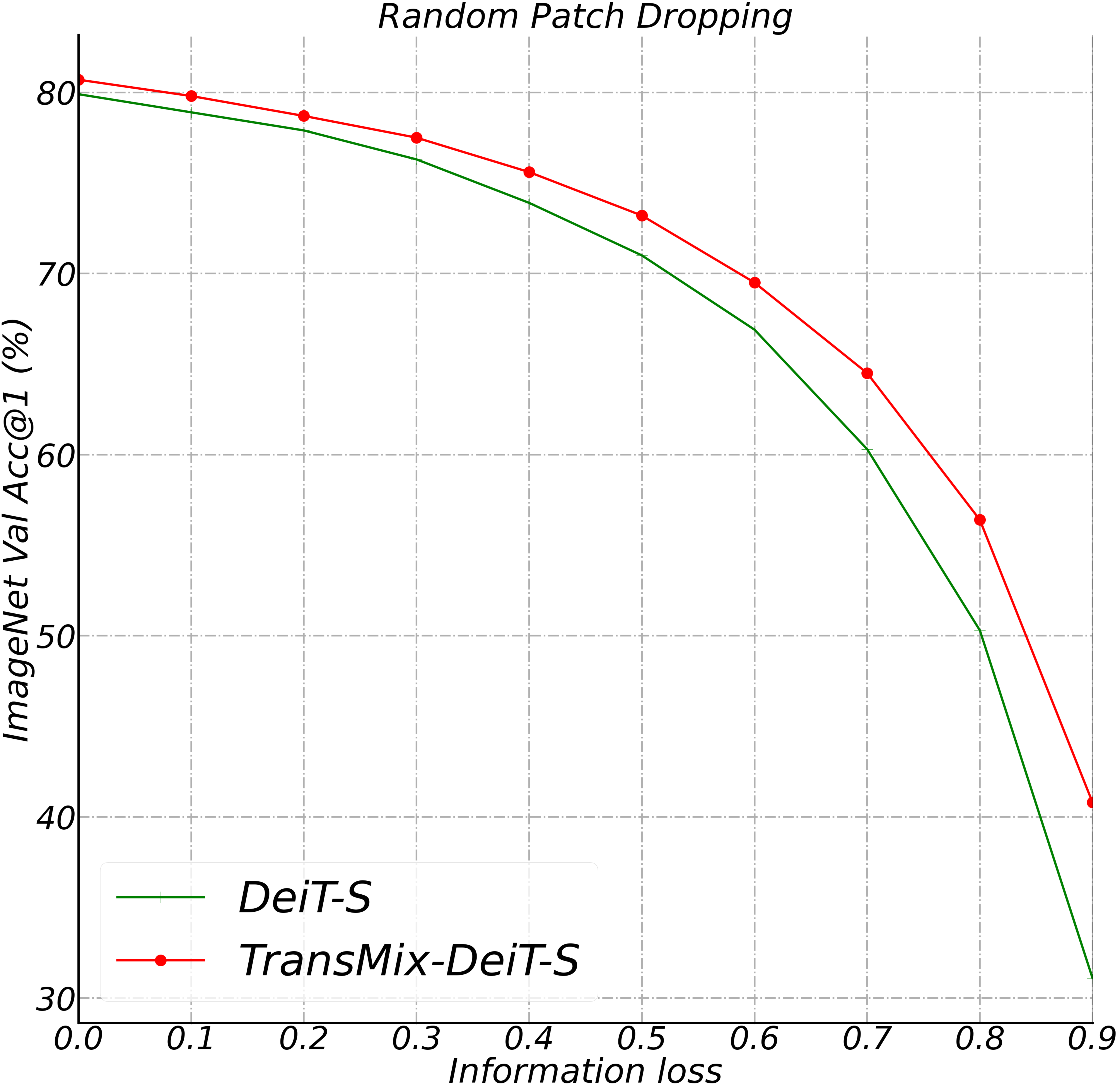}}
\subfigure{\includegraphics[width=0.32\linewidth,height=0.26\linewidth]{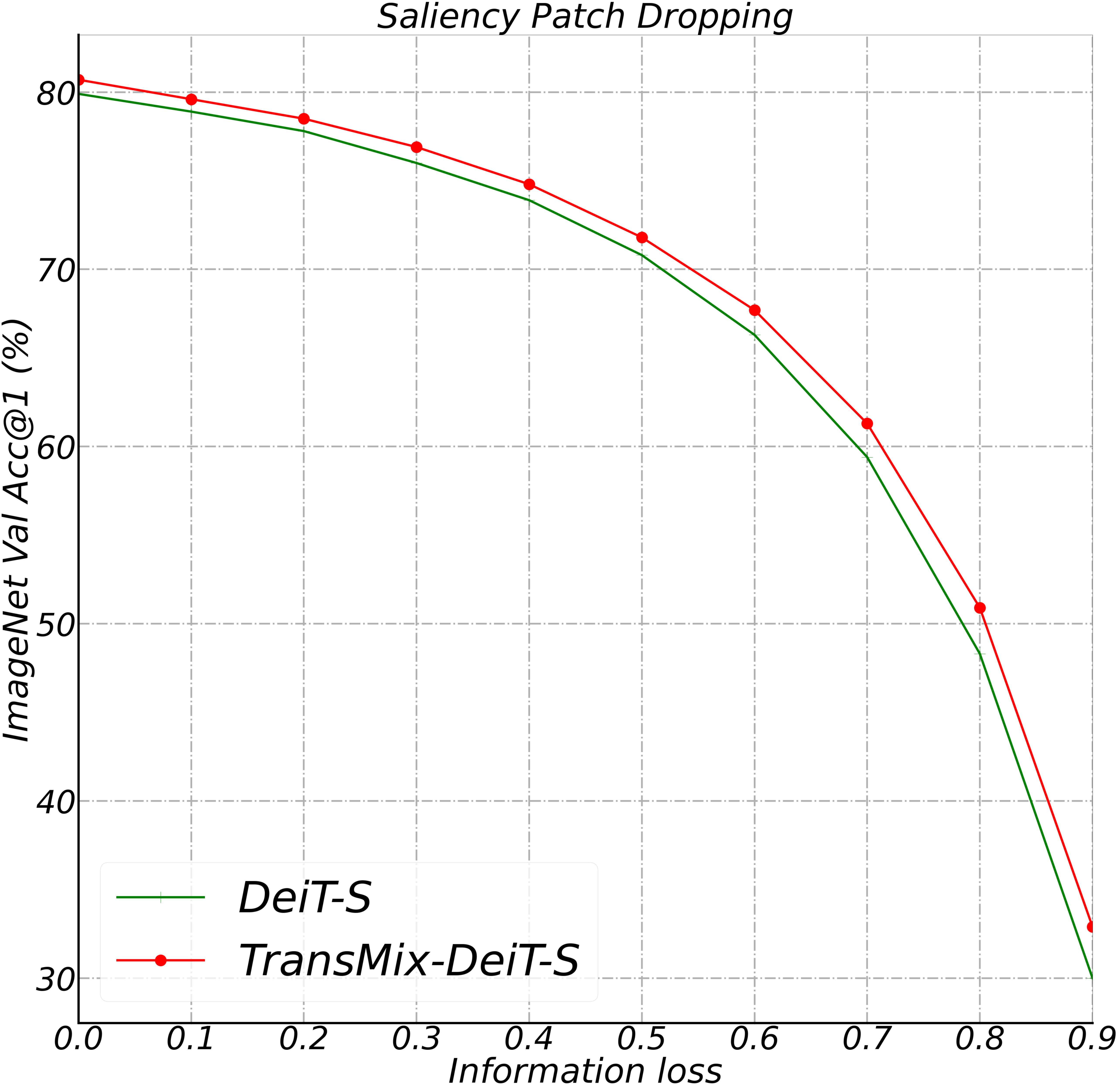}}
\subfigure{\includegraphics[width=0.32\linewidth,height=0.26\linewidth]{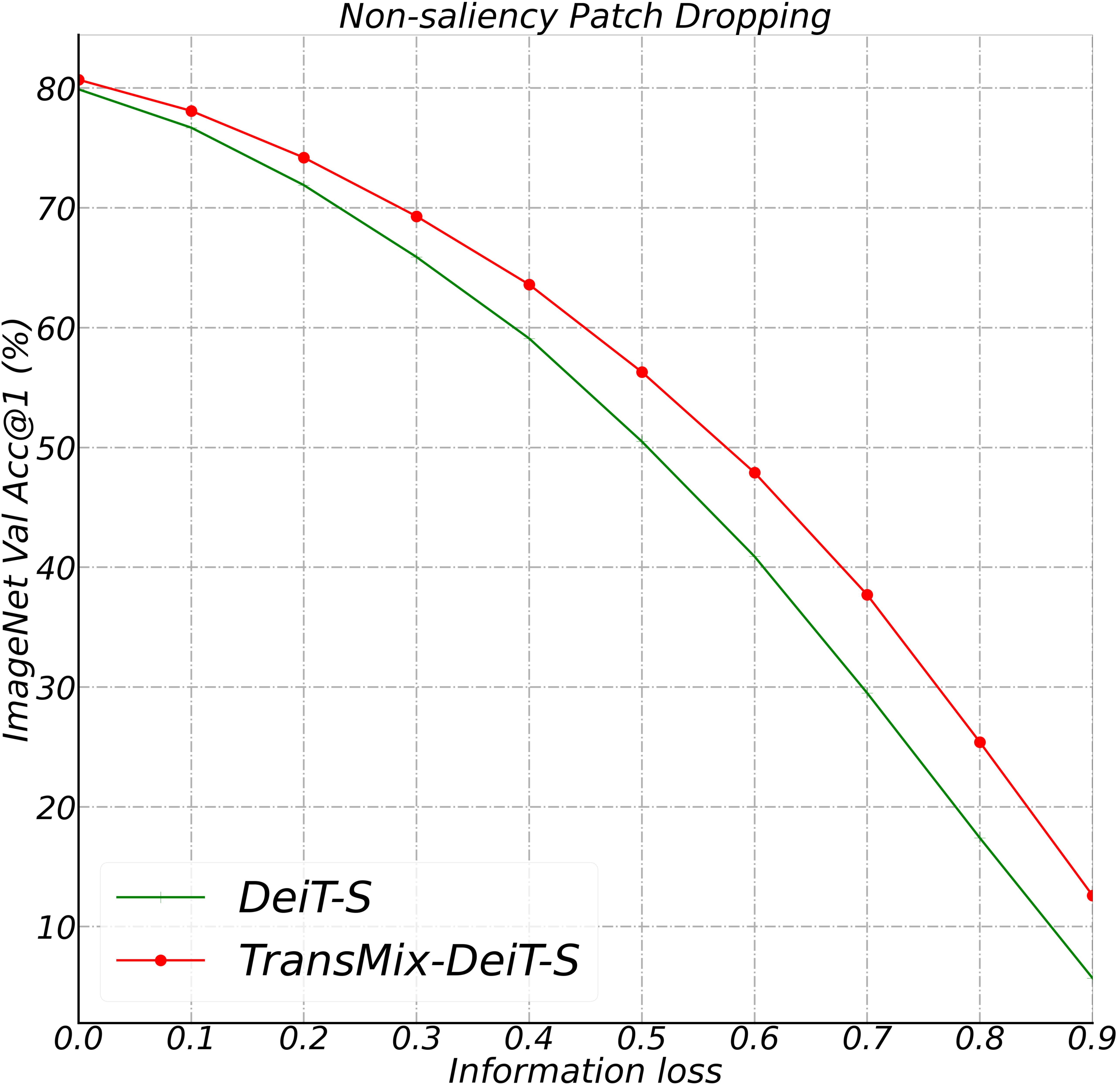}}
\caption{\textbf{Robustness against occlusion.} Model's robustness against occlusion with different information loss ratios is studied. $3$ patch dropping settings: Random Patch Dropping (left), Salient Patch Dropping (middle), and Non-Salient Patch Dropping (right) are considered. }
\label{fig:robust-occ}
\end{figure*}

As shown in Figure.~\ref{fig:robust-occ}, DeiT-S with TransMix outperform vanilla DeiT-S on all occlusion levels especially for extreme occlusion (information loss ratio \textgreater  0.7).

\begin{figure}[!tbp]
\centering
\includegraphics[width=0.95\linewidth]{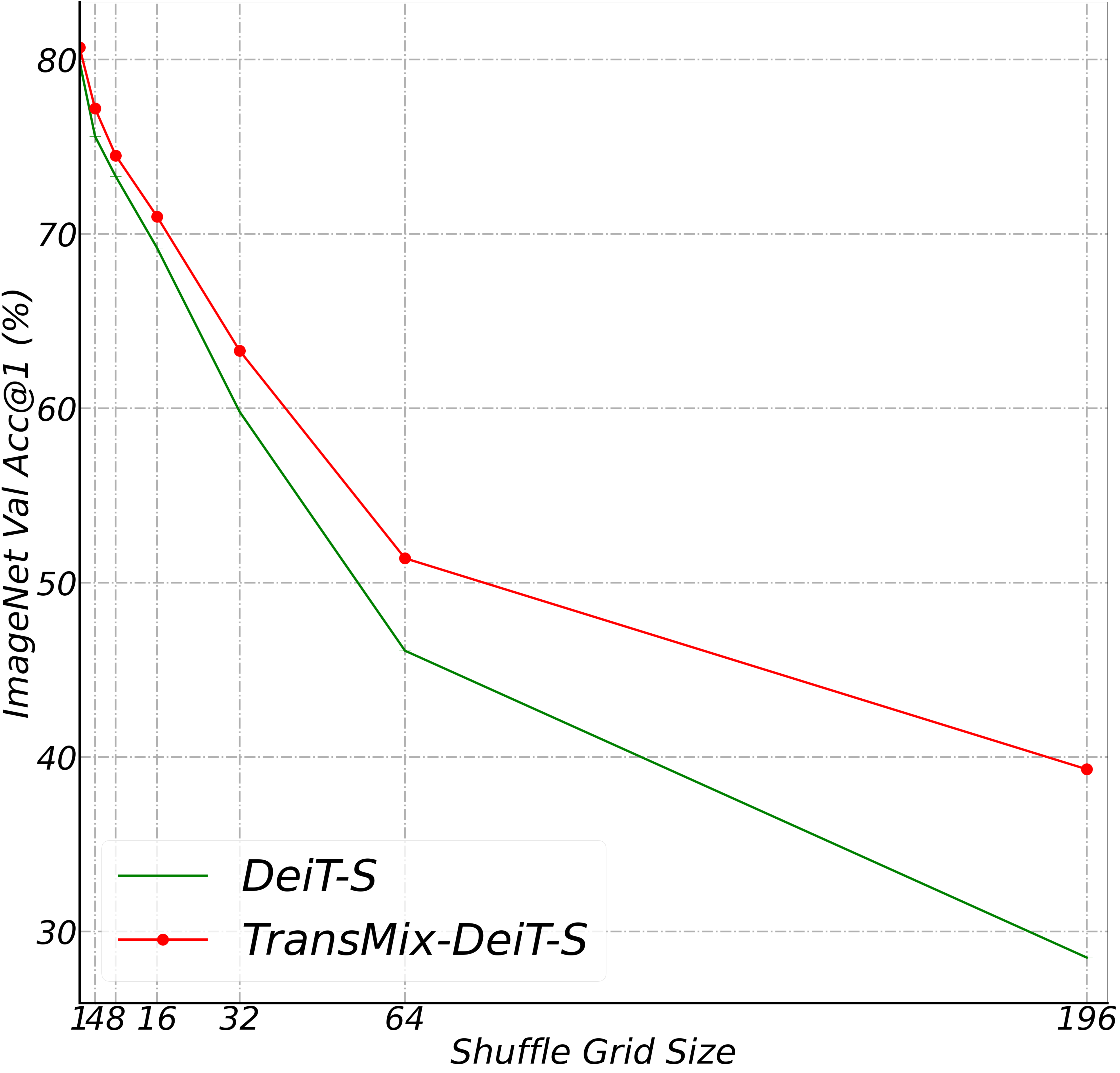}
\caption{\textbf{Robustness against shuffle.} Model's robustness against shuffle with different grid shuffle sizes is studied. (Placeholder)}
\label{fig:robust-shuffle}
\end{figure}

\vspace{1ex}
\noindent\textbf{Sensitivity to Spatial Structure Shuffling}
We study the model’s sensitivity to the spatial structure by shuffling
on input image patches. Specifically, we randomly shuffle the image patches with different grid sizes following~\cite{naseer2021intriguing}. Note that a shuffle grid size of 1 means no shuffle, and a shuffle grid size of 196 means all patch tokens are shuffled. Figure~\ref{fig:robust-shuffle} shows the consistent improvements over baseline, and the accuracy averaged on all shuffled grid sizes for TransMix-DeiT-S and DeiT-S are 62.8\% and 58.4\% respectively.  The superior 4.2\% gain indicates that TransMix enables Transformers rely less on positional embedding to preserve the most informative context for classification.

\vspace{1ex}
\noindent\textbf{Natural Adversarial Example} The ImageNet-A dataset~\cite{hendrycks2021natural} adversarially collects 7500 unmodified, natural but “hard” real-world images, which are drawn from some challenging scenarios (e.g., fog scene and occlusion). The metric for assessing classifiers' robustness to adversarially filtered examples includes the top-1 accuracy, Calibration Error (CalibError)~\cite{kumar2019verified, hendrycks2021natural}, and Area Under the Response Rate Accuracy Curve (AURRA). CalibError judges how classifiers can reliably forecast their accuracy. AURRA is an uncertainty estimation metric introduced in \cite{hendrycks2021natural}. As shown in Table~\ref{tab:ood}, TransMix-trained DeiT-S is superior to vanilla DeiT-S on all metrics. 

\vspace{1ex}
\noindent\textbf{Out-of-distribution Detection}
The ImageNet-O~\cite{hendrycks2021natural} is an adversarial out-of-distribution detection dataset, which  adversarially collects 2000 images from outside ImageNet-1K. The anomalies of unforeseen classes should result in low-confidence predictions. The metric is the area under the precision-recall curve (AUPR)~\cite{hendrycks2021natural}. Table~\ref{tab:ood} indicates that TransMix-trained DeiT-S outperform DeiT-S by 1\% AUPR.

\begin{table}[t]
\tablestyle{2pt}{1.3}
\begin{tabular}{ccccc}
\toprule[1.5pt]
                & \multicolumn{3}{c}{Nat. Adversarial Example}                                               & \multicolumn{1}{c}{Out-of-Dist} \\ \hline
Models          & \multicolumn{1}{c}{Top1-Acc} & \multicolumn{1}{c}{Calib-Error$\downarrow$} & \multicolumn{1}{c}{AURRA} & \multicolumn{1}{c}{AUPR}        \\ \hline
DeiT-S          & 19.1                         & 32.0                            & 23.8                      & 20.9                           \\
\rowcolor[HTML]{F5F5DC}
TransMix-DeiT-S & \textbf{21.1}                         & \textbf{31.2}                            & \textbf{28.8}                      & \textbf{21.9}                           \\ 
\bottomrule[1.5pt]
\end{tabular}
\vspace{2pt}
\caption{Model's robustness against natural adversarial examples on ImageNet-A and out-of-distribution examples on ImageNet-O.}
\label{tab:ood}
\end{table}

\subsection{Mutual Effect of TransMix and Attention}
\label{sec:mutual}
\vspace{1ex}
\noindent\textbf{Will TransMix Benefit Attention?}
To evaluate the quality of attention matrix, we directly threshold the class-token attention  $\mathbf{A}$ from DeiT-S to obtain a binary attention mask (the same with~\cite{caron2021emerging, naseer2021intriguing}] with threshold
0.9) and then conduct two tasks including  (1) \textbf{\textit{Weakly Supervised Automatic Segmentation}} on Pascal VOC 2012 benchmark~\cite{pascal-voc-2012}. (2) \textbf{\textit{Weakly Supervised Object Localization (WOSL)}} on ImageNet-1k validation set~\cite{russakovsky2015imagenet} where the bounding box annotations are only available for evaluation. For task (1), we compute the Jaccard similarity between ground truth and binary attention masks over the PASCAL-VOC12 validation set. For task (2), different from CAM-based methods for CNNs, we directly generate one tight bounding box from the binary attention masks, which is compared with ground-truth bounding box on ImageNet-1k. Both tasks are weakly-supervised since only the class-level ImageNet labels are used for training models (\ie neither bounding box supervision for object localization nor per-pixel supervision for segmentation).
The attention masks generated from TransMix-DeiT-S or vanilla DeiT-S are compared with ground-truth on these two benchmarks. The evaluated scores can quantitatively help us to understand if TransMix has a positive effect on the quality of attention map.

\begin{table}[t]
\tablestyle{2pt}{1.3}
\begin{tabular}{ccc}
\toprule[1.5pt]
             & Segmentation JI (\%) & Localization mIoU (\%) \\ \hline
DeiT-S       & 29.2                 & 34.9                   \\
\rowcolor[HTML]{F5F5DC}
TransMix-DeiT-S & \textbf{29.9}                 & \textbf{44.4}                   \\ 
\bottomrule[1.5pt]
\end{tabular}
\vspace{3pt}
\caption{Quantitative evaluation of the attention map. Segmentation JI denotes the Jaccard index for weakly supervised segmentation on Pascal VOC and Localization mIoU denotes the bounding box mIoU for weakly supervised object localization on ImageNet-1k.} 
\end{table}

\vspace{1ex}
\noindent\textbf{Can Better Attention Nurture TransMix?}
The experiments above prove that TransMix can benefit attention map, and it's natural to ask that can better attention map nurture TransMix in return? We hypothesize that the better attention map is used, the more accurate TransMix adjusts the mixed-target assignment. For example, Dino~\cite{caron2021emerging} confirm that the attention maps obtained from the model via self-supervised training~\cite{caron2021emerging, bao2021beit} retain greater quality. To validate if a better attention map helps TransMix, we design an experiment that replaces the attention map with that generated from a parameter-frozen external model. The external parameter-frozen model can be (1) Dino self-supervised pre-trained DeiT-S (2) Deit-S that is fully-supervised trained on ImageNet-1k. (3)  Deit-S that is fully-supervised trained with a knowledge distillation setting on ImageNet-1k. However, the results shown in Table~\ref{tab:attn_pro} contradict the hypothesis. 

\vspace{1ex}
\noindent\textbf{Intriguing Dynamic Property} With pre-trained Dino as the attention provider, the performance is slightly worse than that of self-serving. Training with attention guidance from a external fully-supervised parameter-frozen DeiT-S, TransMix suffers from a significant drop from 80.7\% to 80.4\% top-1 accuracy, though it is still better than vanilla model's 79.8\%.  This phenomenon can ascribe to the dynamic property of TransMix, meaning that the per-iteration parameter update will dynamically diversify the self-attention for the same input image. In contrast, the parameter-frozen external models statically produce the same self-attention for an image, and thus undermine the regularization capability.

\subsection{Generalizability Study}
\label{sec:swin}
One might be wondering if TransMix can be generalized to those models without the class token such as Swin-Transformer (Swin)~\cite{liu2021swin}. Such models directly apply average pooling onto patch tokens to obtain logits, and therefore how much each patch token contributes to the final prediction is a black-box procedure without class attention $\mathbf{A}$. 

\begin{table}[t]
\tablestyle{2pt}{1.3}
\begin{tabular}{ccccc}
\toprule[1.5pt]
Attn Provider & \cellcolor[HTML]{F5F5DC} Self & Dino & DeiT-pretrained & DeiT-distilled \\ \hline
top-1 Acc     & \cellcolor[HTML]{F5F5DC} 80.7 & 80.6 & 80.4            & 80.4           \\ 
\bottomrule[1.5pt]
\end{tabular}
\vspace{3pt}
\caption{Using external (parameter-frozen) models to generate ttention map as the alternative to original attention map $\mathbf{A}$ used for TransMix.}
\label{tab:attn_pro}
\end{table}
To tackle the aforementioned issue, we develop a Swin variant named as CA-Swin that replaces the last Swin block with a classification attention (CA) block without parameter overhead, which makes it possible to generalize TransMix onto Swin.  Inspired by CaiT~\cite{touvron2021going}, the classification attention block aims at inserting the class token in a plug-and-play manner to those Transformers originally with only patch tokens, and make the classification attention $\mathbf{A}$ accessible. We then compare the Swin-T, CA-Swin-T, TransMix-CA-Swin-T on ImageNet-1k with the same experimental setup in Sec. 4.1. All three models are at the same 28.3M parameters. TransMix-CA-Swin-T and CA-Swin-T have 7\% fewer FLOPs than the baseline Swin-T. The top1 validation accuracy are 81.3\%, 81.6\% and 81.8\% for Swin-T, CA-Swin-T and TransMix-CA-Swin-T, respectively. TransMix on Swin-S improves performance with fewer FLOPs as well. 
This preliminary study empirically proves the generalizability of TransMix.

\begin{table}[t]
\tablestyle{5pt}{1.3}
\begin{tabular}{ccccc}
\toprule[1.5pt]
Models             & Params & FLOPs & top-1 Acc (\%)   \\ \hline
Swin-T~\cite{liu2021swin}              & 28.3M      & 4.5G  & 81.3          \\

CA-Swin-T~\cite{liu2021swin, touvron2021going}           & 28.3M     & 4.2G  & 81.6          \\
\rowcolor[HTML]{F5F5DC}
TransMix-CA-Swin-T     & 28.3M     & 4.2G  & \textbf{81.8}          \\
Swin-S~\cite{liu2021swin}              & 49.6M     & 8.8G  & 83.0          \\
CA-Swin-S~\cite{liu2021swin, touvron2021going}           & 49.6M     & 8.5G  & 82.8          \\
\rowcolor[HTML]{F5F5DC}
TransMix-CA-Swin-S     & 49.6M     & 8.5G  & \textbf{83.2}         \\ 
\bottomrule[1.5pt]
\end{tabular}
\vspace{3pt}
\caption{Generalization to Swin Transformer \cite{liu2021swin} which lacks the class-token. CA denote the class attention block~\cite{touvron2021going}. CA-Swin replaces Swin's last block with a CA block with fewer FLOPs.}
\label{tab:swin}
\end{table}

\vspace{2ex}
\subsection{Comparison with State-of-the-art Mixup Variants}
\label{sec:sota}
\vspace{-5pt}
In this section, we provide the comprehensive comparison with many state-of-the-art mixup vairiants on ImageNet-1k. 
This is the first time that compare these variants on vision Transformer in a fair setting. The implementation details for Mixup variants on top of DeiT-S are provided in the supplementary material. All mentioned models are built upon DeiT training recipe towards a fair comparison. Baseline in Table~\ref{tab:sota} is chosen to be the default DeiT-S framework excluding CutMix in training. Measured on image per second (im/sec), training speed (\ie training throughput) is performed in average of five runs for images at resolution 224$\times$224 under 128 batch size with a Tesla-V100 graphic card, and takes account of data mixup, model forward and backward in train-time.

Table~\ref{tab:sota} shows TransMix significantly outperforms all other Mixup variants. The saliency-based methods (\eg SaliencyMix and Puzzle-Mix) reveal no advantages to vision Transformer, compared to the vanilla CutMix. We analyze that these methods are cumbersomely tuned and face difficulty in  transferring to new architecture. For example, Attentive-CutMix bring not only extra time but also parameter overhead as it introduces an external model to extract saliency map. Puzzle-Mix performs the lowest speed as it forward and backward twice during one training iteration. By contrast, TransMix yields a striking 2.1\% performance advancement with the highest training throughput and no parameter-overhead.

\begin{table}[t]
\tablestyle{0.8pt}{1.3}
\begin{tabular}{ccccc}
\toprule[1.5pt]
Method & Backbone & Params & \begin{tabular}[c]{@{}c@{}} Speed \\ (im/sec) \end{tabular} &  top-1 Acc (\%) \\
\hline
Baseline & \multirow{6}{*}{DeiT-S} &  22M & 322 &   78.6 \\
CutMix~\cite{yun2019cutmix} & &  22M & 322 &  79.8 \gain{1.2} \\
Attentive-CutMix~\cite{walawalkar2020attentive} &  & 46M & 239 &    77.5 \loss{1.1} \\
SaliencyMix~\cite{uddin2020saliencymix} &  &  22M & 314 &   79.2 \gain{0.6} \\
Puzzle-Mix~\cite{kim2020puzzle} &  &  22M & 139 &   79.8 \gain{1.2} \\
\rowcolor[HTML]{F5F5DC}
TransMix & &  22M & 322 &   \textbf{80.7} \gain{2.1} \\
\bottomrule[1.5pt]
\end{tabular}
\vspace{3pt}
\caption{ \textbf{Top1-accuracy, training speed (im/sec) and number of parameters} comparison with state-of-the-art Mixup variants on ImageNet-1k. All listed models are built upon DeiT training recipe for fair comparison. Training speed (im/sec) takes account of data mixup, model forward and backward in train-time.}

\label{tab:sota}
\end{table}

\vspace{1ex}
\noindent\textbf{Ablation Study} Unlike suprisingly 8 hyper-parameters in PuzzleMix, our proposed TransMix exists very clean and introduces almost no hyper-parameter. Still, we conduct ablation study for TransMix regarding the attention map generation in the supplementary material, which shows that the default is the best.

\begin{figure}[!tbp]
\centering
\includegraphics[width=\linewidth]{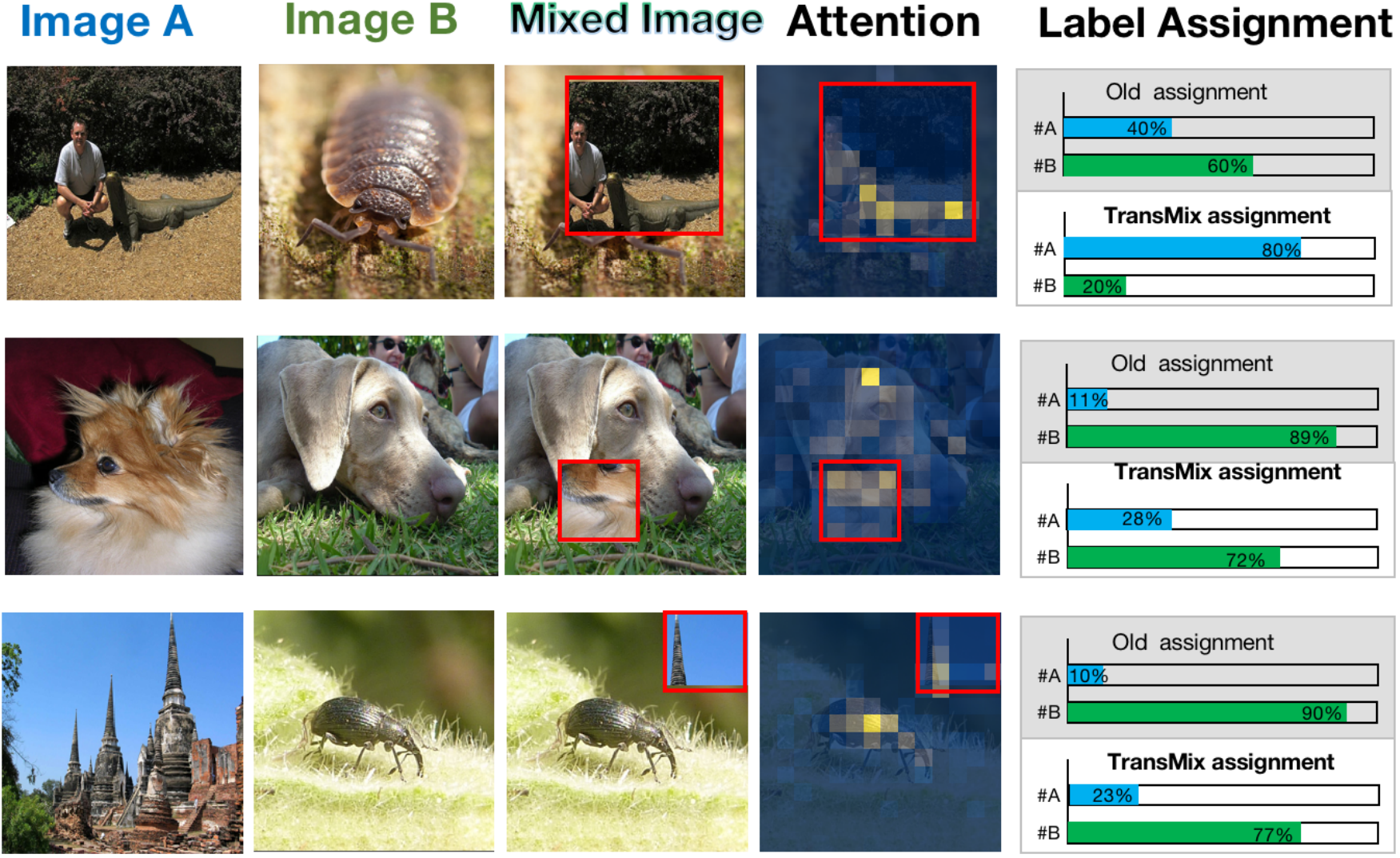}
\caption{The visualization including image A, image B, mixed image, attention map obtained from XCiT-L when input mixed image, and corresponding label assignments. The label assignments include both the old area-ratio assignment and new TransMix assignment. }
\label{fig:viz}
\end{figure}

\vspace{1ex}
\noindent\textbf{Visualization} We provide the visualization of TransMix as shown in Figure~\ref{fig:viz}. For instance, the first row illustrate that the old area-based label assignment is counter-intuitive as image A's foreground is occluded by image B's patch, and TransMix corrects the label assignment via Transformer attention. TransMix is able to lift the label weight if the discriminative fine-grained attribute appears (\eg Pomeranian dog's cheek and eyes in the second row). 

\section{Conclusion}
 
In this paper, we present TransMix, a simple yet effective data augmentation technique that assigns Mixup labels with attentional guidance for Vision Transformers. TransMix naturally exploits Transformer's attention map to assign the confidence for the mixed-target, and lifts the top-1 accuracy on ImageNet by 0.9\% for both DeiT-S and a large variant XCiT-L. Extensive experiments are conducted to verify the effectiveness, transferability, robustness and generalizability of TransMix on totally 10 benchmarks.

\vspace{2ex}
\noindent\textbf{Limitations}
Since we are the first work that pushes an extra mile for the Mixup-based methods towards augmenting vision Transformers, we indeed have limitations as follows: 

\noindent(1) TransMix can not handle well with those backbones without class token, as it strongly relies on the class attention. This limitation can be mitigated in Section~\ref{sec:swin} at the cost of architecture modification.

\noindent(2) TransMix requires the attention map to be spatially aligned with the input, which indicates that it may not be compatible with deformable-based Transformer (e.g. PS-ViT~\cite{yue2021vision}, DeformDETR~\cite{zhu2020deformable}). In the future, this problem can be potentially solved by calibrating attention map to the input spatial location by leveraging deformed offset grid.

\noindent(3) As the cropped patch with sharp rectangle boundary is strikingly distinguishable to background, the Transformer may be naturally curious about the cropped patch and then attending to the patch, resulting in a base attentional weight no matter if the patch contains useful information. This phenomenon will also occur in previous saliency-based methods since the cropped patch edge will enhance the first-/second-order feature statistics.

\newpage
\vspace{1ex}
\noindent\textbf{Acknowledgement}
We would like to thank Xiaoyu Yue, Huiyu Wang, Qihang Yu and Chen Wei for their feedbacks on the paper. This work is done while the first two authors intern at Bytedance Inc. Shuyang Sun and Philip Torr are supported by the ERC grant ERC-2012-AdG 321162-HELIOS, EPSRC grant Seebibyte EP/M013774/1 and EPSRC/MURI grant EP/N019474/1. We would also like to thank the Royal Academy of Engineering and FiveAI.

\appendix

\section{Additional Experimental Details}
\subsection{Implementation Details of Compared Mixup Variants}
The comparison with state-of-the-art Mixup variants is conducted in Section 4.6. We explain the implementation details here. The official implementations of Mixup variants are mainly based on the backbone of ResNet-50, and we apply their methods into training DeiT-S.

\vspace{1ex}
\noindent\textbf{Baseline}
Baseline in Table~\ref{tab:sota} is chosen to be the default DeiT-S framework excluding CutMix in training.

\vspace{1ex}
\noindent\textbf{Attentive-CutMix}
Attentive-CutMix is implemented based on the unofficial pytorch repository \footnote{https://github.com/xden2331/attentive\_cutmix}. Attentive-CutMix contains an affiliated model (\ie ResNet-50) for saliency map extraction and a backbone model for image classification. 

\vspace{1ex}
\noindent\textbf{SaliencyMix}
Saliency-Mix is implemented based on the official pytorch codebase \footnote{https://github.com/afm-shahab-uddin/SaliencyMix}. SaliencyMix uses third-party library opencv to extract the saliency map with 
\begin{lstlisting}[basicstyle=\footnotesize]
cv2.saliency.StaticSaliencyFineGrained_create()
\end{lstlisting}

\vspace{1ex}
\noindent\textbf{Puzzle-Mix}
Puzzle-Mix is implemented following the official pytorch codebase \footnote{https://github.com/snu-mllab/PuzzleMix}. Puzzle-Mix forwards and backwards the model twice to detect object saliency by computing the gradients of the neural network following~\cite{simonyan2013deep}.

\section{Additional Results}
\vspace{1ex}
\noindent\textbf{Ablation Study}
The class attention $\mathbf{A}$ can obtained from any Tranformer Block in ViTs. Due to the global receptive field, the class attention would not have big difference across blocks~\cite{raghu2021vision, dosovitskiy2020image}. We first study the effect of  attention matrix generated in different depth $d$ for DeiT-S. Then we follow~\cite{abnar-quantifying, Chefer_2021_CVPR} to compute the attention rollout, which aggregate the attention matrices from all blocks by matrix multiplications. According to the results, we found that the default setting with $d=12$  performs the best. Notably, the total number of Transformer block with class token is varying in different vision Transformers (\eg 24 for XCiT, 2 for CaiT, 12 for DeiT). Particularly, PVT designs hierarchical Transformer blocks with 4 different resolution scales, and therefore an extra downsample step is a must if using early scale attention matrices. Hence, using the attention from the last Transformer block as default can not only avoid finding a optimal $d$ exhaustingly but also be compatible for all ViT variants.
\begin{table}[t]
\tablestyle{2pt}{1.3}
\begin{tabular}{ccccccccc}
\hline
$d$ &  6    & 8  & 10  & 12  & rollout \\
\hline
top-1 Acc & 80.3 &  80.3 & 80.4 & 80.7 &  80.4 \\
\hline
\end{tabular}
\vspace{3pt}
\caption{\textbf{Ablation study} on attention generation. Attention matrix used for TransMix is output from the $d$-th block of DeiT-S. Following~\cite{abnar-quantifying, Chefer_2021_CVPR}, rollout applies matrix multiplication across all 12 blocks' attention matrices.}
\end{table}

\vspace{1ex}
\noindent\textbf{Mixup variants on CNN and ViT}
We also attach the official results of some Mixup variants with the backbone of CNN. Results on the ResNet-50 backbone are borrowed from~\cite{kim2020puzzle}. All models are trained for 300 epochs towards fair comparison. As backbone, DeiT-S has similar number of parameters to ResNet-50.
Table~\ref{tab:sota-cnn} shows that SaliencyMix and Puzzle-Mix only improve over CutMix by at most 0.2\% on ResNet-50 and show no advancement over CutMix on DeiT-S. 
\begin{table}[t]
\tablestyle{0.8pt}{1.3}
\begin{tabular}{ccccc}
\toprule[1.5pt]
Method & Backbone & Params &  top-1 Acc (\%) \\
\hline
Baseline & \multirow{4}{*}{ResNet-50} &  25M  &   76.3 \\
CutMix~\cite{yun2019cutmix} & &  25M  &  78.6  \\
SaliencyMix~\cite{uddin2020saliencymix} &  &  25M  &   78.7  \\
Puzzle-Mix~\cite{kim2020puzzle} &  &  25M  & 78.8 \\

\hline
Baseline & \multirow{6}{*}{DeiT-S} &  22M  &   78.6 \\
CutMix~\cite{yun2019cutmix} & &  22M  &  79.8 \\
Attentive-CutMix~\cite{walawalkar2020attentive} &  & 46M  &    77.5 \\
SaliencyMix~\cite{uddin2020saliencymix} &  &  22M  &   79.2  \\
Puzzle-Mix~\cite{kim2020puzzle} &  &  22M  &   79.8  \\
TransMix & &  22M &   80.7 \\
\bottomrule[1.5pt]
\end{tabular}
\vspace{3pt}
\caption{  Comparison with state-of-the-art Mixup variants with the backbone of either ViT or CNN on ImageNet-1k. All listed models are trained for 300 epochs towards fair comparison. ResNet-50 results are borrowed from the paper~\cite{uddin2020saliencymix}.}

\label{tab:sota-cnn}
\end{table}


{\small
\bibliographystyle{plainnat}
\bibliography{egbib.bib}
}

\end{document}